\documentclass{article}

\usepackage[english]{babel}

\usepackage[letterpaper,top=2cm,bottom=2cm,left=3cm,right=3cm,marginparwidth=1.75cm]{geometry}

\usepackage{amsmath}
\usepackage{graphicx}
\usepackage[colorlinks=true, allcolors=blue]{hyperref}

\usepackage{multicol}
\usepackage{multirow}

\providecommand{\keywords}[1]
{
  \small	
  \textbf{\textit{Keywords---}} #1
}

\makeatletter
\newenvironment{tablehere}
{\def\@captype{table}}
{}

\newenvironment{figurehere}
{\def\@captype{figure}}
{}
\makeatother

\title{Active Learning-Based Optimization of Scientific Experimental Design}
\author{Ruoyu Wang\\School of Computer Science \& Engineering\\South China University of Technology\\Guangzhou, Guangdong, 510006, China\\lannywong2000@163.com}
\date{ }

\begin{document}
\maketitle

\begin{abstract}
Active learning (AL) is a machine learning algorithm that can achieve greater accuracy with fewer labeled training instances, for having the ability to ask oracles to label the most valuable unlabeled data chosen iteratively and heuristically by query strategies. Scientific experiments nowadays, though becoming increasingly automated, are still suffering from human involvement in the designing process and the exhaustive search in the experimental space. This article performs a retrospective study on a drug response dataset using the proposed AL scheme comprised of the matrix factorization method of alternating least square (ALS) and deep neural networks (DNN). This article also proposes an AL query strategy based on expected loss minimization. As a result, the retrospective study demonstrates that scientific experimental design, instead of being manually set, can be optimized by AL, and the proposed query strategy ELM sampling shows better experimental performance than other ones such as random sampling and uncertainty sampling. 
\end{abstract}

\keywords{Machine Learning, Active Learning, Scientific Experiment Design, Alternating Least Square, Deep Neural Network}

\begin{multicols}{2}

\section{Introduction}

Scientific experiments have become progressively reliant on automated machines overtime to cope with large datasets and high concurrency. For instance, Hafner et al.\cite{hafner2017designing} propose a mostly automated pipeline for drug response experiments implementation and results quantification. However, the remained human involvement in the process of experimental design has a high tendency to generate errors that are difficult to tackle. The most conventional replacement for manual experimental design is an exhaustive one. But without heuristic sampling methods, the exhaustive exploration in the experiment space tends to lead to excessive consumption of experimental resources and manpower.

Some non-exhaustive methods were proposed to achieve experimental design automation, such as random sampling and orthogonal experimental design\cite{ruijiang2010study}. However, since the two methods are based on stochastic distribution and predetermined orthogonal table respectively, both of them fail to take the properties of the experimental subjects into consideration and therefore their performance varies in real-life experimental practices.

Accordingly, an active method is needed to implement automated experimental design adaptively based on the properties of the experimental subjects. And active learning (AL) is a machine learning algorithm that can achieve greater accuracy with fewer labeled training instances, for having the ability to ask oracles to label the most valuable unlabeled data chosen iteratively by query strategies\cite{settles2009active}.

In the regard of above, this article performs a retrospective study on a dataset produced by the mentioned pipeline and discusses the feasibility of adaptive design of scientific experiments using AL strategy. Insights and inspiration can be found in this article for scientists to harness the strength of AL to avoid human involvement in experimental design, heuristically achieve experimental design optimization and train satisfying predictive models with relatively fewer physical experiments.

\section{Dataset}

\subsection{Dataset Description}

The dataset used in this article is Breast Cancer Profiling Project, Drug Sensitivity 1\cite{_2018}. This dataset,  produced from the mentioned drug response experiment pipeline\cite{hafner2017designing} of Hafner’s as shown in Figure \ref{fig:1}, contains the exhausted experimental results of measuring the sensitivities of 35 breast cancer cell lines to 34 small molecule perturbagens. Every combination of cell and molecule is treated with 9 doses of different concentrations of small molecule perturbagens respectively. Therefore, the dataset contains 10710 instances in total.

\begin{figurehere}
\centering
\includegraphics[width=0.5\textwidth]{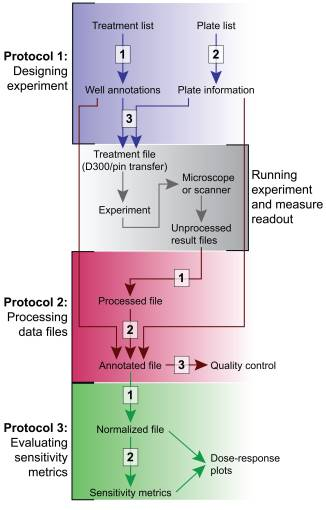}
\caption{\label{fig:1}Hafner’s Pipeline for experimental design and analysis}
\end{figurehere}

The independent and dependent variables of the dataset are described as follows. The independents variables are Cell HMS LINCS ID, Cell Name, Small Molecule HMS LINCS ID, Small Molecule Name, Small Mol Concentration (uM), Primary Target and Pathway. The dependent variables are two measurements of sensitivity that are Mean Normalized Growth Rate Inhibition Value (GR) and Increased Fraction Dead (IFD).

\[GR = 2^{\frac{\log_2\left(\frac{x(c)}{x_0}\right)}{\log_2\left(\frac{x_{ctrl}}{x_0}\right)}}\cite{hafner2016growth}\]

Where $x(c)$, $x_0$ and $x_{ctrl}$ is the mean of the measured live cell counts after a given treatment, from the day 0 untreated plate grown in parallel until the time of treatment, and of the control wells for all technical replicates respectively.

\[IFD = fd(c) - fd_{ctrl}\]

Where $fd(c)$ is the mean fraction of dead cells in the wells from a given treatment and $fd_{ctrl}$ is the mean fraction of dead cells in the control wells across all technical replicates.

\subsection{Properties of Variables}

High interdependencies exist among the independent variables. For example, Cell HMS LINCS ID and Cell Name have a one-to-one mapping relationship, and Primary Target and Pathway are both determined by each certain combination of cell and molecule. Thus, the independent variables of the dataset can be restated as Cell Type, Molecule Type and Molecule Concentration (uM).

Notably, the explicit physical or chemical properties of cells and molecules are not given in the dataset.

The independent variables measuring sensitivity are continuous. Thus, a regression model with regression metrics, such as rooted mean square error, is needed to perform prediction work.

Scientists are also interested in the classification accuracy that whether certain molecule accelerates or restrains the growth of certain cell. Obviously, a GR value greater than 1 or a negative IFD value indicates a positive effect of certain molecule on the growth of certain cell. On the contrary, a GR value less than 1 or a positive IFD value indicates a negative one. In the following experiments, the GR values are typically reduced by 1 so that both the independent variables have a binary classification boundary of 0.

\subsection{Dataset Redefinition}

As previously discussed, every combination of cell and molecule are treated with 9 different concentrations of molecule respectively. However, the 9 concentrations are not entirely the same as each other. For the convenience of doing retrospective AL experiments (i.e. to make sure that the sensitivity result is known for every possible combination of cell, molecule and concentration of molecule), the problem space will be restricted to a subset of all concentration values in which every concentration value has correspondent results of all combinations of cell and molecule. As it turns out, this subset of concentration values has a size of 4.

Therefore, the problem space will be redefined to one containing 4760 samples. Although the size of the dataset is halved, it is still worth studying since one single experiment takes three days to complete in the  physical world in a serialized manner\cite{_2018}, and a better experimental design can avoid excessive consumption of resources and can also produce more accurate predictive models.

In the following experiments, all the proposed methods will be performed on the 4 subdatasets of different concentration values separately. Particularly, the training curves of the subdataset with a molecule concentration of 0.01uM are shown to illustrate the traits of the algorithms. And the arithmetic mean of all the 4 subdatasets’ results is calculated when referring to average.

\end{multicols}

\begin{figurehere}
\centering
\includegraphics[width=1\textwidth]{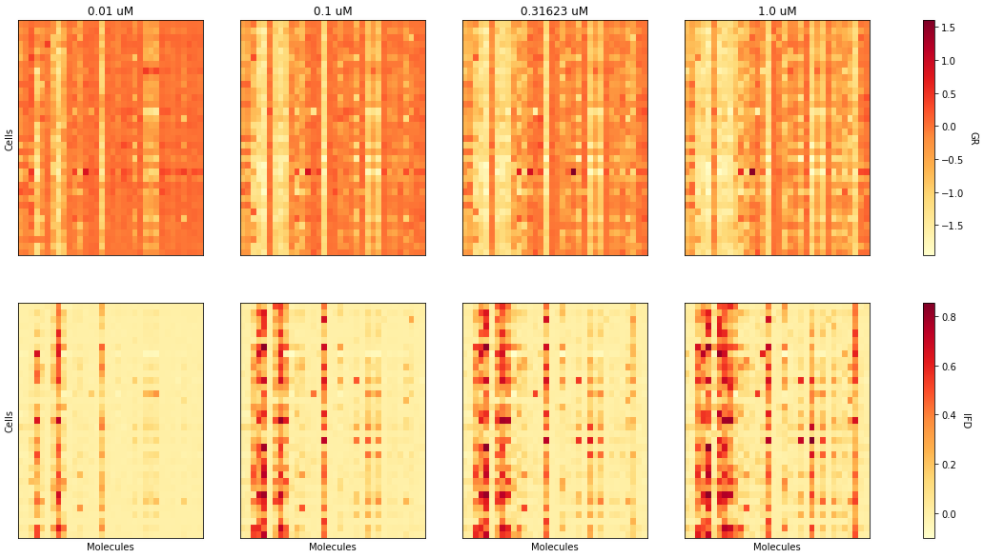}
\caption{\label{fig:2}Heatmap visualization of the redefined dataset}
\end{figurehere}

\begin{multicols}{2}

\section{Predictive Models}

The difficulty of building predictive models for the dataset lies in that the independent variables do not represent the explicit features of cells and molecules, therefore cannot be used to train predictive models directly. In this section, a matrix factorization method called Alternating Least Square (ALS) will be used to produce the latent features of cells and molecules, then an ALS improvement method with the combination of Deep Neural Network (DNN) will be proposed.

\subsection{Metrics}

The metrics for the predictive models are loss and accuracy. Rooted mean square error (RMSE) is used to measure the loss. And accuracy is calculated as the percentage of predicted sensitivity values that are on the same side of decision boundary (0 in this article) as the ground truth sensitivity values.

\[RMSE = \sqrt{\frac{1}{n}\sum_{i=1}^n\left(\hat{y}_i-y_i\right)^2}\]

$n$ is the number of samples, $\hat{y}_i$ is the predicted value and $y_i$ is the ground truth for sample $i$.
 
\subsection{Alternating Least Square}

\subsubsection{ALS Prerequisites}

ALS\cite{hu2008collaborative} is a collaborative filtering algorithm in the field of recommender systems using matrix factorization. Specifically, given a sparse user-item real matrix recording user ratings on the items, the aim is to decompose the user-item matrix into a dot product of a user embedding matrix and an item embedding matrix such that the squared error between the dot product and the user-item matrix is minimized. The loss function is shown below where $i$, $j$, $d$, $x$, $w$, $y$ and $r$ is the $i^{th}$ of $m$ users, $j^{th}$ of $n$ items, number of dimensions of embedding vector, $m\times{}d$ user embedding matrix, $d\times{}n$ item embedding matrix, $m\times{}n$ user-item matrix and $m\times{}n$ 0-1 matrix recording rated positions respectively.

\[loss = \frac{1}{2}\sum_{(i,j)\in r(i,j)=1}\left(\sum_{l=1}^dx_{il}w_{lj}-y_{ij}\right)^2\]

One can then deduce the partial derivatives of loss with respect to $x$ and $w$.

\[\frac{\partial loss}{\partial x_{ik}} = \sum_{j\in r(i,j)=1}\left(\sum_{l=1}^dx_{il}w_{lj}-y_{ij}\right)w_{kj}\]

\[\frac{\partial loss}{\partial w_{kj}} = \sum_{i\in r(i,j)=1}\left(\sum_{l=1}^dx_{il}w_{lj}-y_{ij}\right)x_{ik}\]

Finally, the gradient decent method can be used to train the model with learning rate $\alpha$.

\[x = x-\alpha\Delta x = x-\alpha r\cdot\left(xw-y\right)w^T\]

\[w = w-\alpha\Delta w = w-\alpha x^T\left[\left(xw-y\right)\cdot r\right]\]

This method is called alternating because the user matrix and the item matrix are trained alternately while the other one is fixed.

\subsubsection{ALS Experiment Settings}

The number of dimensions of the embedding vector $d$ is intuitively set to 5. The learning rate $\alpha$ is 0.01. An ALS model is trained for every cell-molecule matrix and a 10-fold cross-validation is performed with 400 epochs of training for every ALS model.

\subsubsection{ALS Results Analysis}

\begin{figurehere}
\centering
\includegraphics[width=0.5\textwidth]{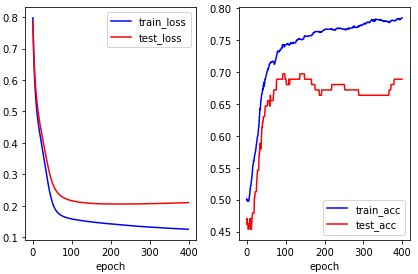}
\caption{\label{fig:3}ALS training curve of GR with molecule concentration of 0.01uM}
\end{figurehere}

\begin{figurehere}
\centering
\includegraphics[width=0.5\textwidth]{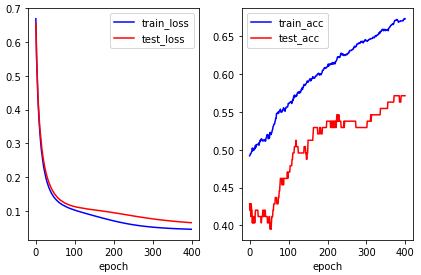}
\caption{\label{fig:4}ALS training curve of IFD with molecule concentration of 0.01uM}
\end{figurehere}

The ALS training curves of GR and IFD with molecule concentration of 0.01 uM are shown in Figure \ref{fig:3} and Figure \ref{fig:4} and the average test loss and average test accuracy are shown in Table \ref{table:1}. From Figure \ref{fig:3} and Figure \ref{fig:4}, it is revealed that:

The training loss, RMSE, converges at a relatively satisfying value of around 0.15 for GR and around 0.05 for IFD after 100 to 200 epochs of training.

The phenomenon of overfitting is quite obvious especially as shown in Figure \ref{fig:3}, where the test loss only converges at around 0.2 and lacks behind the training loss by roughly 0.1.

The model reduces loss at the expense of accuracy. After 100 to 200 epochs of training, the test loss drops at an extremely slow rate while the test accuracy is stable or even reduced. This reduction of test accuracy can significantly impair the practical performance of the predictive model and thus should be addressed carefully.

\subsection{Alternating Least Square with Deep Learning (ALSDL)}

\subsubsection{ALS Improvement with DL}

An ALS improvement using Deep Learning (DL) method is proposed to solve the problems above. After several training epochs of the ALS algorithm, the latent features of cell and molecule under certain concentration can be achieved by user (cell) and item (molecule) matrixes. Then the aligned features of cell and molecule are fed to a Fully Connected Neural Network (FCNN) to predict the sensitivity value.

A new loss function is proposed with RMSE reducing a penalty term for incorrect classification, where $\beta$ is a weight parameter and n is the size of the training set, so that the model has the tendency to correctly classify the sensitivity value. Specifically when the RMSE is fixed, the loss increases as more sensitivity values are incorrectly classified thus producing a negative product of prediction and ground truth, vice versa.

\[loss = RMSE - \beta\frac{1}{n}\sum_{i=1}^nsign\left(\hat{y}_i,y_i\right)\]

\[sign(x) = \begin{cases}
1, & x>0 \\
0, & x=0 \\
-1, & x<0 
\end{cases}\]

A more general form of the proposed loss function is shown below, where the predictive model is tackling a k-class classification problem with classification boundaries $c_1,c_2,......,c_k$. As a brief explanation, only a prediction falls in the same interval between two adjacent classification boundaries as the ground truth will produce a positive penalty term, otherwise, negative.

\[loss = RMSE - \beta\frac{1}{n}\sum_{i=1}^nsign\left(\sum_{j=1}^ksign\left(\left(\hat{y}_i-c_j\right)\left(y_i-c_j\right)-k+1\right)\right)\]

\subsubsection{ALSDL Experiment Settings}

The parameters of ALS in ALSDL are the same as in the former experiment. The FCNN has three hidden layers containing 20, 10, 5 neural cells respectively and an output layer of 1 neural cell. The activation functions between hidden layers are all tanh and the optimizer for back propagation is rmsprop. The weight parameter $\beta$ of the loss function is 0.1. An ALSDL model is trained for every cell-molecule matrix and a 10-fold cross-validation is performed with 200 epochs of ALS training and another 200 epochs for FCNN training for every ALSDL model.

\subsubsection{ALSDL Result Analysis}

\begin{figurehere}
\centering
\includegraphics[width=0.5\textwidth]{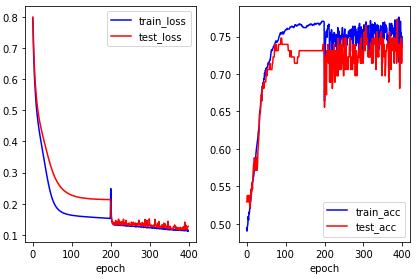}
\caption{\label{fig:5}ALSDL training curve of GR with molecule concentration of 0.01uM}
\end{figurehere}

\begin{figurehere}
\centering
\includegraphics[width=0.5\textwidth]{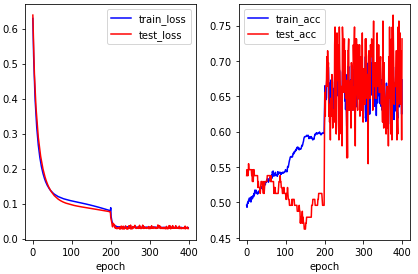}
\caption{\label{fig:6}ALSDL training curve of IFD with molecule concentration of 0.01uM}
\end{figurehere}

The ALSDL training curves of GR and IFD with molecule concentration of 0.01 uM are shown in Figure \ref{fig:5} and Figure \ref{fig:6} and the results are promising in three aspects comparing with ALS.
The loss, RMSE, continues to converge at a lower value of around 0.1 for GR and around 0.02 for IFD with a significant improvement compared with ALS convergence value at 200 epochs of training.

The phenomenon of overfitting is less intense as the curves of training loss and test loss almost overlap during the 200 training epochs of DL.

With the restriction of the penalty term for incorrect classification, the accuracy does not decrease as loss reduces at the DL training stage. Instead, for both GR and IFD, the test accuracy is not significantly worse if not better than the training accuracy.

\begin{tablehere}
\centering
\resizebox{\linewidth}{!}{
\begin{tabular}{c|c|c|c|c}
\multirow{2}*{ } & \multicolumn{2}{c|}{Average Test Loss} & \multicolumn{2}{c}{Average Test Accuracy} \\ \cline{2-5}
                 & ALS & ALSDL & ALS & ALSDL \\ \hline
GR               & 0.250860068 & 0.160126785 & 0.854432773 & 0.872478992 \\ \hline
IFD              & 0.0097235223 & 0.054959413 & 0.748781513 & 0.818277311 \\
\end{tabular}}
\caption{\label{table:1}ALS and ALSDL average test loss and average test accuracy}
\end{tablehere}

Table \ref{table:1} shows that, for both GR and IFD, the average test loss is reduced and the average test accuracy is increased by ALSDL compared with ALS.

\section{Active Learning Implementation}

With the predictive model ALSDL proposed and its effectiveness examined, the AL algorithm is to be implemented on the model to form the AL scheme as a whole. However, incompatibility occurs when common classification-based query strategies, such as the uncertainty sampling, are performed on the regression problem. In this section, a query strategy based on expected loss minimization and suitable for regression problem will be proposed followed by comparative experiment and result analysis.

\subsection{Active Learning Prerequisites}

Active learning (AL) is a machine learning algorithm that can train models to achieve greater accuracy or less error with relatively fewer labeled training instances. Figure \ref{fig:7} illustrates the procedure of pool-based AL where all the unlabeled data are maintained as a pool and are queried by query strategies till the pool is empty. Carefully designed query strategies will actively choose the most valuable unlabeled data for oracles to label, thus a more representative training set and a better-trained model can be obtained. The most common query strategies include uncertainty sampling, least confident and margin sampling\cite{settles2012active}.

\begin{figurehere}
\centering
\includegraphics[width=0.5\textwidth]{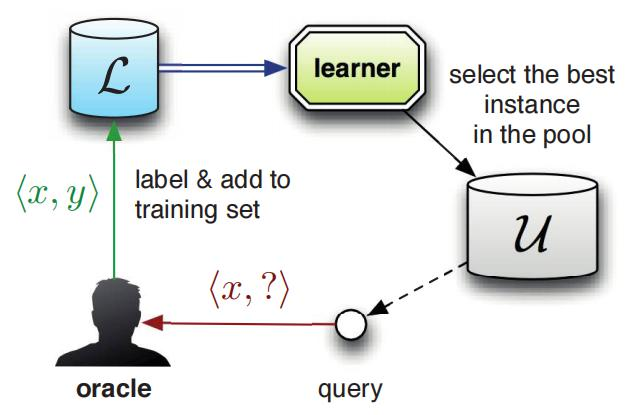}
\caption{\label{fig:7}Pool-Based Active Learning}
\end{figurehere}

\subsection{Query Strategy based on Expected Loss Minimization}

To implement an AL algorithm on the predictive model, a query strategy based on the minimization of the expected loss (ELM) of the model is proposed.

The core idea of ELM query strategy is to choose the unlabeled sample $\left(x^+,y^+\right)$ from the unlabeled dataset $U$ such that when $\left(x^+,y^+\right)$ is removed from $U$ and added to the current model $M$’s training set $D$ to train a new model $M^+$, the new model produces the least test loss. The test loss is measured with the combination of $D$ and $U$, i.e. the whole dataset.

\[x_{ELM} = \mathop{\arg\min}\limits_{x^+\in U }loss\left(M^+,D\cup U\right)\]

However, since the ground truth of $y^+$ and $U$’s label values are unknown, the expected new model $\hat{M}^+$ is trained with the current model $M$’s prediction of $\left(x^+,\hat{y}^+\right)$ and training set $D$, then the expected test loss of $\hat{M}^+$ is evaluated with $M$’s predictions of unlabeled data $\hat{U}$ in substitution of $U$.

\[x_{ELM} = \mathop{\arg\min}\limits_{x^+\in U }loss\left(\hat{M}^+,D\cup \hat{U}\right)\]

Multiple samples can also be obtained in one query by sorting the unlabeled samples in ascending order according to their expected loss then selecting the top ones.

\subsection{Proposed Active Learning Scheme and Experiment Settings}

The procedures of the proposed AL scheme are as follows.

Initialization: Randomly choose $n_{init}$ samples to be labeled as the training set $D$. $n_{query}=0$.

Training: Train a new ALSDL predictive model on $D$.

Query: If $n_{query}>n_{max_query}$, terminate the scheme. Otherwise, select $n_{per_query}$ samples  to be added to $D$ from the unlabeled dataset using ELM query strategy, $n_{query} = n_{query}+1$, then return to procedure Training.

Both $n_{init}$ and $n_{per_query}$ are set to 40 and $n_{max_query}$ is 8. The ALSDL model has the same settings as in the former experiment. The RMSE loss and accuracy of the ALSDL model evaluated on the full dataset are recorded after every training procedure. The RMSE loss  is used for new model evaluation in the ELM query strategy and is evaluated after 200 epochs of ALS training to speed up the query procedure.

In comparison, three other sampling methods (query strategies) are implemented to substitute ELM query strategy in the scheme. Orderly sampling resembles a manual experimental design that queries user-item matrix in a row or column order. Random sampling queries the unlabeled pool stochastically. Uncertainty sampling chooses the unseen samples with sensitivity values closest to the classification boundary.

\subsection{Active Learning Result Analysis}

The proposed AL scheme training curves of GR and IFD with molecule concentration of 0.01uM are shown respectively as Figure \ref{fig:8} and Figure \ref{fig:9}. After 8 times of queries, with a total of 360 examples learned, the proposed ELM sampling method reaches the highest accuracy and the least loss for both GR and IFD. Compared with other query strategies, the proposed expected loss minimization query strategy constructs better predictive models more efficiently for the following aspects.

\begin{figurehere}
\centering
\includegraphics[width=0.5\textwidth]{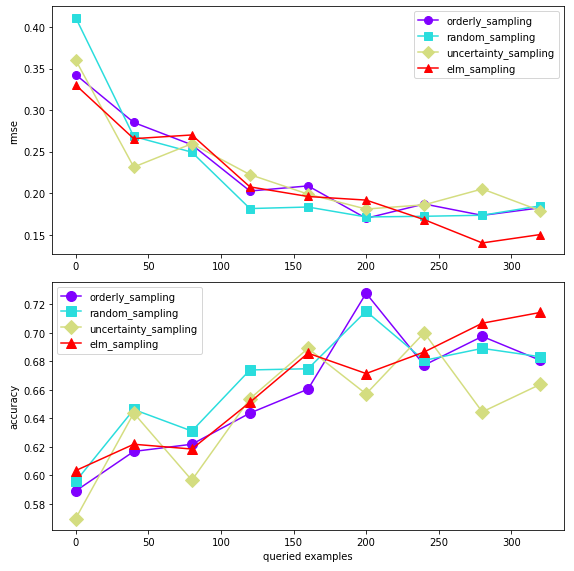}
\caption{\label{fig:8}Active learning training curve of GR with molecule concentration of 0.01uM}
\end{figurehere}

\begin{figurehere}
\centering
\includegraphics[width=0.5\textwidth]{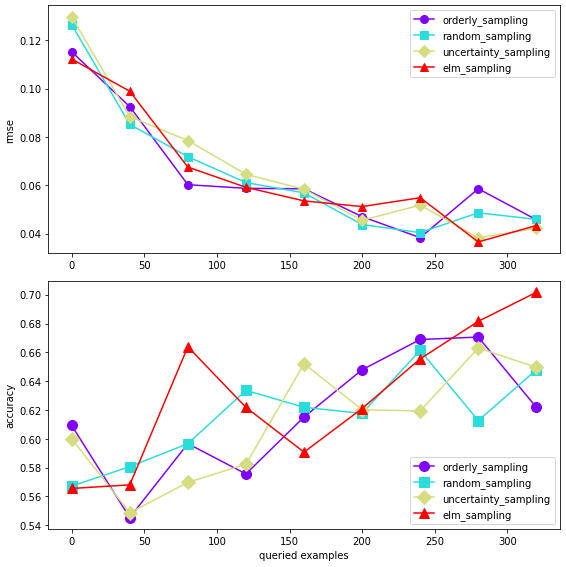}
\caption{\label{fig:9}Active learning training curve of IFD with molecule concentration of 0.01uM}
\end{figurehere}

The rate of convergence of ELM sampling is more rapid. With little percent of total experiment space covered, the performance of ELM sampling falls behind other sampling methods in the beginning. Then ELM soon catches up and outperforms other sampling methods as the expected test loss becoming more and more accurate.
ELM sampling, whose loss function always encourages the training of a more accurate model, is more heuristically designed than orderly and random sampling. It also prevails over uncertainty sampling because the latter depends on the classification boundary that is constantly shifting with the result of the ALS model’s training.
With the same amount of examples queried, ELM sampling results in a predictive model of less loss and higher accuracy because the queried samples are among the most valuable ones.
Therefore, the optimized design of the experiment of cells’ sensitivity to molecules can be achieved by the proposed AL scheme with the query strategy of expected loss minimization.

\section{Conclusion}

This article proposed an AL scheme that combines ALS, DNN and a query strategy variant for conducting scientific experiments design without prior knowledge of the properties of experimental subjects. After performing a retrospective study on a dataset of scientific experiment results, it was demonstrated that the optimization of experimental design that builds better predictive models more efficiently can be automatically reached by the AL scheme and the proposed ELM query strategy outperforms several others.

Due to the constraint of computational power, all the hyperparameters in this article are not heuristically tuned but manually set based on the optima of several trials. Similarly, the FCNN setting used in this article is not necessarily optimal and it requires further investigation. However, the performance should not vary significantly as the hyperparameters shift since the whole scheme is built on mature and robust algorithms.

The work in this article can be an inspiration for optimizing the design of automated scientific experiments, especially whose consumption of physical experimental resources is tremendous and the knowledge of experimental subjects’ properties is limited, such as drug response experiments in the field of biomedicine. In the future, more datasets would be tested and more experiments would be conducted on the proposed AL scheme to further improve its generality.

\section*{Acknowledgement}

This article would not have been finished without the valuable reference materials and lectures that I received from the professor, Dr. Robert F. Murphy of Carnegie Mellon University, whose insightful guidance and enthusiastic encouragement gave me much help in the course when I was shaping this article hence I want to show my deepest gratitude to him.

I would also avail myself of this opportunity to extend my sincere thanks to the teachers from whose teaching and instruction I obtained during my undergraduate education in the School of Computer Science \& Engineering, South China University of Technology.

Last but not least, I am much indebted to my family, without whose unconditional affection and consistent support this article could not have appeared in its final form.

\end{multicols}

\end{document}